\documentclass[sigconf]{acmart}
\usepackage{graphicx}
\usepackage{amsmath}
\usepackage{booktabs}
\usepackage{multirow}

\AtBeginDocument{%
  \providecommand\BibTeX{{%
    \normalfont B\kern-0.5em{\scshape i\kern-0.25em b}\kern-0.8em\TeX}}}

\begin{document}

\title{Graph Attention Transformer Network for Multi-Label Image Classification}

\author{Jin Yuan}
\affiliation{%
\institution{Southeast University}
\city{Nanjing}
\country{China}
}
\email{yuanjin@seu.edu.cn}

\author{Shikai Chen}
\affiliation{%
\institution{Southeast University}
\city{Nanjing}
\country{China}
}
\email{skchen@seu.edu.cn}

\author{Yao Zhang}
\affiliation{%
\institution{University of Chinese Academy of Sciences}
\city{Beijing}
\country{China}
}
\email{zhangyao215@mails.ucas.ac.cn}

\author{Zhongchao Shi}
\affiliation{%
\institution{Lenovo Research}
\city{Beijing}
\country{China}
}
\email{shizc2@lenovo.com}

\author{Xin Geng}
\affiliation{%
\institution{Southeast University}
\city{Nanjing}
\country{China}
}
\email{xgeng@seu.edu.cn}

\author{Jianping Fan}
\affiliation{%
\institution{Lenovo Research}
\city{Beijing}
\country{China}
}
\email{jfan1@lenovo.com}

\author{Yong Rui}
\affiliation{%
\institution{Lenovo Research}
\city{Beijing}
\country{China}
}
\email{yongrui@lenovo.com}


\begin{abstract}
Multi-label classification aims to recognize multiple objects or attributes from images. However, it is challenging to learn from proper label graphs to effectively characterize such inter-label correlations or dependencies. Current methods often use the co-occurrence probability of labels based on the training set as the adjacency matrix to model this correlation, which is greatly limited by the dataset and affects the model's generalization ability. In this paper, we propose a Graph Attention Transformer Network (GATN), a general framework for multi-label image classification that can effectively mine complex inter-label relationships. First, we use the cosine similarity based on the label word embedding as the initial correlation matrix, which can represent rich semantic information. Subsequently, we design the graph attention transformer layer to transfer this adjacency matrix to adapt to the current domain. Our extensive experiments have demonstrated that our proposed methods can achieve highly competitive performance on three datasets. \textcolor{red}{https://github.com/a791702141/GATN}.
\end{abstract}



\keywords{multi-label classification, attention mechanism, graph neural network, transformer}


\maketitle


\section{Introduction}
Multi-label image classification is a basic visual task, which aims to recognize multiple objects or attributes in an image. It has wide real-world applications, like human expression classification \cite{zhao2016deep, li2016human, zhuang2018multi, chen2020label}, social tag recommendation \cite{nam2019learning, vu2020privacy} and medical image-aided diagnosis \cite{ge2018chest}. The most critical challenge for multi-label image classification is to learn the inter-label relationships and their dependencies in various images. 

Motivated by the powerful ability of graph neural networks (GNN) in processing relation information, current works \cite{chen2019multi, wang2020multi, NguyenAAAI2021} often exploited GNN to model the correlation between categories, thereby achieving highly competitive results for multi-label classification. As a core of GNN, relation adjacency matrix represent the relational strength of different nodes (categories), which motivate us to solve multi-label classification by enhancing the expressive ability of relation adjacency matrix. We consider the issue faced by adjacency matrix as two parts: (1) How to define the relation between different categories. (2) How to transfer this knowledge into specific task. 

For these two problems, conventional methods \cite{chen2019multi, wang2020multi, NguyenAAAI2021} often adopted a fixed co-occurrences matrix as a relation adjacency matrix. However, relations include more than just appearing together, they also include similar appearances (e.g. \emph{bear} and \emph{teddy bear}), similar functions (e.g. \emph{airplane} and \emph{truck}) and etc, which are difficult to learn 
from image datasets that need to be manually annotated. On the other hand, billion-level corpus contains a wealth of knowledge and information, which is often used for generating expressive word embedding. Therefore, a natural idea is to leverage pre-trained word embeddings to build adjacency relation matrix instead of a co-occurrence one.

\begin{figure}
	\centering
	\includegraphics[width=\columnwidth]{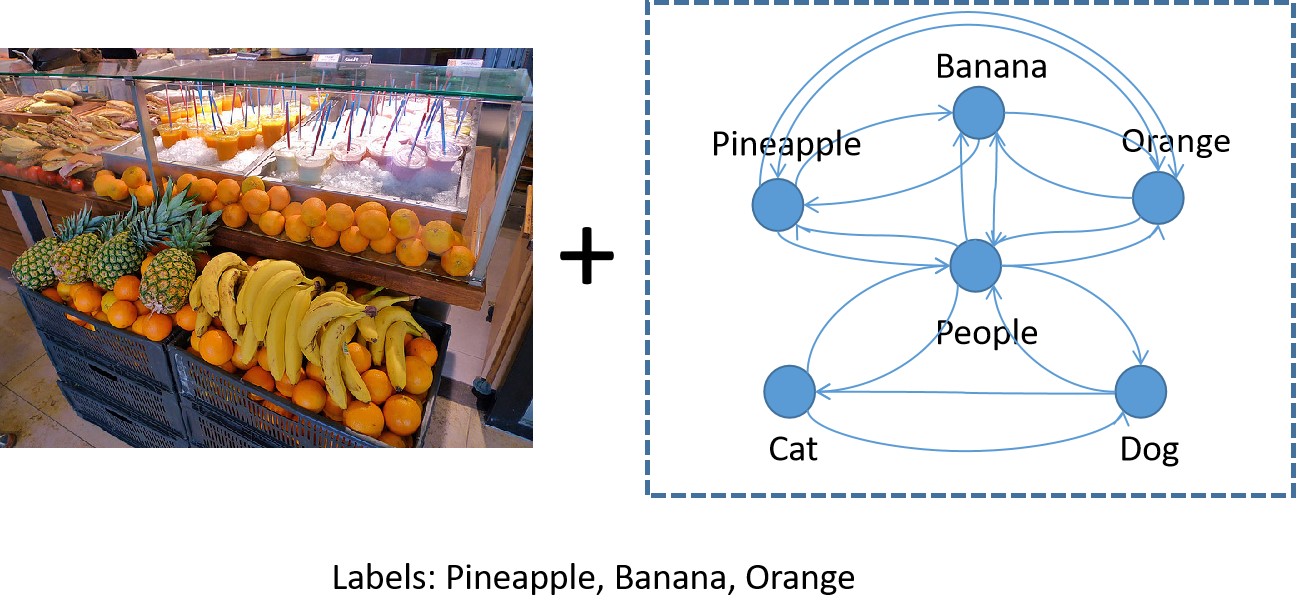}
	\caption{This figure shows the general framework for GNN-based methods. The left part is the input image, and the part on the right is the topological diagram of the label relationship, usually represented by the adjacency matrix.}
	\label{fig11}
\end{figure}

Given a relation matrix with rich information, how to adapt it to a specific task is another crucial issue. Inspired by the remarkable ability of the self-attention mechanism in processing word vector relationships, we find that the self-attention mechanism is essentially a graph-like structure, which is helpful in extracting the inter-label relationships. Based on this observation, we propose a novel graph attention transformer layer to transfer the correlation matrix by explicit attention block to mine its internal sub-graph structure. Generally, our proposed graph attention transformer network can effectively borrow information from the adjacency matrix and then adapt it into the specific task. 

Our contributions are as follows:
\begin{itemize}

\item We propose a novel end-to-end GATN framework to gain helpful information from node representation by self-attention branches, which can more accurately identify the meta-path of the graph.
\item We initialize the node correlation matrix in the graph by the label node embedding, which has richer semantic information than the previous co-occurrence probability method.
\item We evaluate our proposed model in several datasets, and the results demonstrate the highly competitive performance compared with the existing methods.
\end{itemize}

\section{Related Works}

\subsection{Multi-label Classification}
As a fundamental task in computer vision, multi-label classification has drawn much attention in past years. Some early methods solved this task by dividing the task into multiple binary classification tasks \cite{read2011classifier, quevedo2012multilabel, liu2015on}. With the rise of convolutional neural networks (CNN), some works \cite{yang2016exploit, gao2017deep, nath2019single} used the CNN-based method to complete this task. These methods usually used binary cross-entropy and sigmoid function to learn image features, thereby neglecting the information between the labels. Therefore, some approaches adopted a CNN-RNN based architecture \cite{wang2016cnn, liu2017semantic, chen2017order, yazici2020orderless} to predict labels in order. Wang et al. \cite{wang2016cnn} emphasized the importance of predicting simple labels first and then predicting other labels with the previous prediction result information. To break the limitation of predicting labels according to frequency, Chen et al. \cite{chen2017order} exploited an LSTM network to generate label order, making the prediction order learning parameters. Yazici et al. \cite{yazici2020orderless} proposed a predicted label alignment to avoid the further misclassification caused by the misalignment of the label order.

\subsection{Graph Neural Network}
The graph neural network (GNN) has proven to be helpful in a variety of tasks \cite{perozzi2014deepwalk, tang2015line, grover2016node2vec, lee2018multi, veli2018graph}, like solving node classification in social networks or assisting other tasks to extract connections between nodes. As variants of graph neural networks, many networks have different advantages in processing graph structure data. Graph convolutional network \cite{bruna2014spectral, kipf2016semi} was introduced as a scalable method for semi-supervised learning on graph-structured data, and it can effectively extract the features between graph nodes. On the other hand, the graph transformer network \cite{yun2019graph} was designed to solve the problem of the heterogeneous graph by mining the structure of different sub-graphs to generate new meta-paths.

Due to the powerful ability of graph neural networks in dealing with the data relationship for graph-structured data, some works \cite{chen2019multi, wang2020multi, NguyenAAAI2021} solve multi-label classification based on image features and label topological structure. Chen et al. \cite{chen2019multi} exploited GCN to gain label features by inputting label word embedding and co-occurrence matrix as node representation and adjacency matrix, respectively. They achieved state-of-the-art performance at that time by blending the label features and image features learned by GCNs and brought the trend of studying multi-label image classification by graph neural network. Wang et al. \cite{wang2020multi} proposed a label graph superimposing framework to enhance the graph-based network further. To mine more sub-graph structures, Nguyen et al. \cite{NguyenAAAI2021} introduced a modular architecture with a graph transformer network to optimize the GCN-based method by exploring multiple sub-graph structures.

\subsection{Attention Mechanism}

Attention mechanism was designed to mine the relevance between different entities in language and visual task \cite{vaswani2017attention}. Due to its great success in a relational understanding, attention mechanism began to be widely used in Natural Language Processing (NLP) to understand the relationship between different words \cite{bahdanau2015neural, devlin2018bert}. As a core of the transformer, the attention mechanism regained its brilliance in the Computer Vision (CV) \cite{chen2020generative, dosovitskiy2021an, Liu2021survey} and its extraction of entity relationships was very competitive. For the multi-label classification task, building label dependencies is a critical problem, which motivates us to take advantage of the attention mechanism on label correlation learning.

\begin{figure*}
	\centering
	\includegraphics[width=\textwidth]{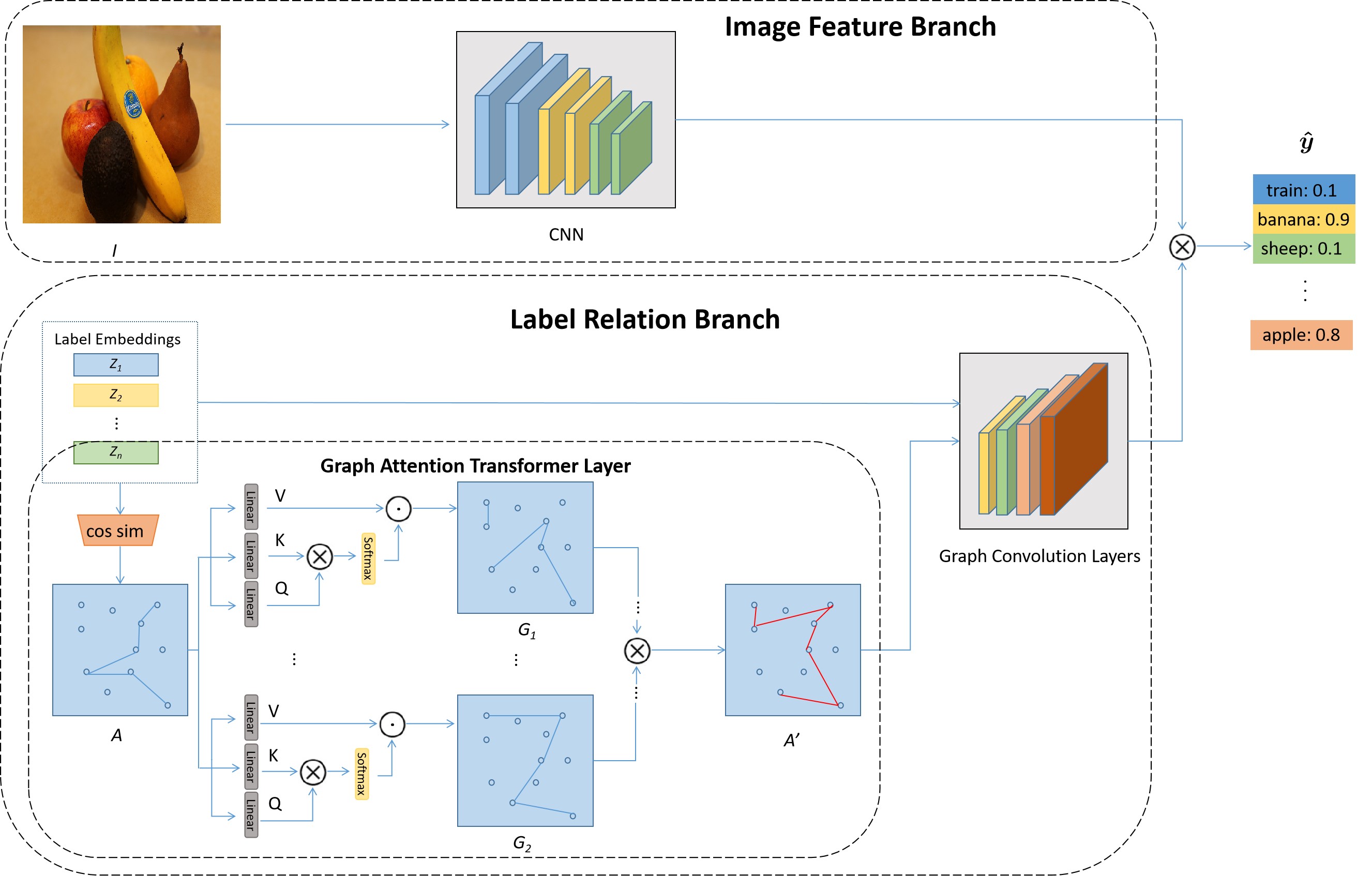}
	\caption{The overview of our method. In the label relation branch, we first initialize the adjacency matrix by cosine similarity of the label node embedding. Then we adopt a graph attention transformer layer to mine sub-graph structures.}
	\label{fig1}
\end{figure*}

\section{Method}

Our proposed framework has two branches. The image feature branch uses a common pre-trained model to extract image features, and the label relation branch aims to learn the relationship among labels. In the label relation branch, Our proposed Graph Attention Transformer Network focuses on learning a richer label node correlation information.
 
Firstly, we propose a novel generation scheme of correlation matrix based on the distance of node embedding. Then we exploit the graph attention transformer layer to obtain the weight matrix for transferring the generated matrix. The overview is shown in Fig.~\ref{fig1}.

\subsection{Generation of Correlation Matrix}
For $ n $ categories multi-label classification, the adjacency matrix $\boldsymbol{A} \in \mathbb{R}^{n \times n}$ is expected to include rich semantic relational information of labels. In graph neural network, the node representations $\boldsymbol{Z} \in \mathbb{R}^{n \times d}$ usually comes from word embeddings trained in an unsupervised large-scale corpus, which contain rich semantic information. Therefore, we use node embeddings to construct the correlation matrix.

Let $\boldsymbol{Z}_{i} $ denote embedding vector of $i$th label of length $ d $, and $ \boldsymbol{R}_{i,j} $ represent the relation value between $i$th label and $j$th label, we obtain $ \boldsymbol{R}_{i,j} $ by calculating the cosine similarity of the embedding vectors:

\begin{equation}
\boldsymbol{R}_{i,j} = \frac{\sum_{k=1}^{d} \boldsymbol{Z}_{i,k} \times \boldsymbol{Z}_{j,k}}{\sqrt{\sum_{k=1}^{d}\left(\boldsymbol{Z}_{i,k}\right)^{2}} \times \sqrt{\sum_{k=1}^{d}\left(\boldsymbol{Z}_{j,k}\right)^{2}}}.
\end{equation}

The obtained matrix $ \boldsymbol{R} $ is symmetric, $ \boldsymbol{R}_{i,j} $ is always equal to $ \boldsymbol{R}_{j,i} $, which means that the relationship between any two categories is consistent.

To avoid some noisy edges and over-smoothing problems, we use binary and re-weighted strategies like to filter and re-weight values \cite{chen2019multi}. With threshold $\tau$ and weight parameter $ p $, we can get final correlation matrix $\boldsymbol{A} $:

\begin{equation}
	\boldsymbol{R}_{i,j}^{\prime}=\left\{\begin{array}{ll}
		0, & \text { if } \boldsymbol{R}_{i,j}< \tau \\
		1, & \text { if } \boldsymbol{R}_{i,j} \geq \tau 
	\end{array}\right.
,
\end{equation}

\begin{equation}
	\boldsymbol{A}_{i,j}=\left\{\begin{array}{ll}
		p / \sum_{j=1 \atop i \neq j}^{n} \boldsymbol{R}_{i,j}^{\prime}, & \text { if } i \neq j \\
		1-p, & \text { if } i=j
	\end{array}\right.
	.
\end{equation}

\subsection{Graph Attention Transformer Layer}

We design a novel graph transformer layer by focusing on node representation. Thus, our proposed graph attention transformer layer can transform the adjacency matrix into a new graph structure and explore new multi-hop paths. Define each sub-graph selected from adjacency matrix $ \boldsymbol{A} $ as $ \boldsymbol{G}_{j} $. Each $ \boldsymbol{G}_{j} $ can be regarded as a transformation of $ \boldsymbol{A} $ with some new structure information. On the other hand, the self-attention mechanism has a powerful ability in structural transformation, which has led a great success for transformers in computer vision. Therefore, we use self-attention mechanism based branches to obtain different sub-graphs.

We use standard multi-head self-attention layer similar to \cite{vaswani2017attention}. For input correlation matrix $\boldsymbol{A}$, we can get $ \boldsymbol{Q}_{i} $, $ \boldsymbol{K}_{i} $, $ \boldsymbol{V}_{i} $ for $ i $th head through different linear layers:

\begin{equation}
\begin{aligned}
	& [\boldsymbol{Q}_{i}, \boldsymbol{K}_{i}, \boldsymbol{V}_{i}] = \boldsymbol{A} [\boldsymbol{W}_{i}^{Q}, \boldsymbol{W}_{i}^{K}, \boldsymbol{W}_{i}^{V}] \\
	 & \quad \boldsymbol{W}_{i}^{Q}, \boldsymbol{W}_{i}^{K}, \boldsymbol{W}_{i}^{V} \in \mathbb{R}^{n \times D_{h}},
\end{aligned}
\end{equation}
where $ D_{h} $ is the hidden size of $ \boldsymbol{Q}_{i} $, $ \boldsymbol{K}_{i} $ and $ \boldsymbol{V}_{i} $. Then the attention layer for $i$th head can be writen as:

\begin{equation}
	\text{Attention}(\boldsymbol{Q}_{i}, \boldsymbol{K}_{i}, \boldsymbol{V}_{i})=\operatorname{softmax}\left(\frac{\boldsymbol{Q}_{i} \boldsymbol{K}_{i}^{T}}{\sqrt{D_{h}}}\right)\boldsymbol{V}_{i}.
\end{equation}

For a $ h $-head attention layer, we can get a sub-graph adjacency matrix $ \boldsymbol{G}_{j} \in \mathbb{R}^{n \times n} $:

\begin{equation}
\begin{aligned}
	& \boldsymbol{G}_{j}=\text{Concat}(\text{Attention}(\boldsymbol{Q}_{1}, \boldsymbol{K}_{1}, \boldsymbol{V}_{1}), \\
	& \ldots, \text{Attention}(\boldsymbol{Q}_{h}, \boldsymbol{K}_{h}, \boldsymbol{V}_{h}))\boldsymbol{W^{O}}.
\end{aligned}
\end{equation}

Since we want each $ \boldsymbol{G}_{j} $ to learn a different sub-graph structure, each $ \boldsymbol{G}_{j} $ uses attention layers of non-shared parameters. For $ k $ generated sub-graphs adjacency matrix $ \boldsymbol{G}_{1}, \boldsymbol{G}_{2}, ..., \boldsymbol{G}_{k} $, we intend to integrate them into the transformed adjacency matrix $ \boldsymbol{A}^{\prime} $. The most straightforward solution is to adopt matrix multiplication:

\begin{equation}
	\boldsymbol{A}^{\prime} = \prod_{j=1}^{k} \boldsymbol{G}_{j}.
\end{equation}

In general, the graph attention transformer layer can be regarded as a transfer function used to adapt the adjacency matrix to any feature extraction layer:

\begin{equation}
	\boldsymbol{A}^{\prime} = f(\boldsymbol{A}).
\end{equation}

\subsection{Graph Attention Transformer Networks}

With the transformed adjacency matrix $ \boldsymbol{A}^{\prime} $ and node embedding $ \boldsymbol{Z} $, we adopt graph convolutional network \cite{kipf2016semi} to learn useful representations for label node. Let $ \boldsymbol{H}^{(l)}$ be the label node representations of the $l$th layer in GCNs, the forward propagation can be writen as:

\begin{equation}
	\boldsymbol{H}^{(l+1)}=\sigma\left(\boldsymbol{\tilde{D}}^{-\frac{1}{2}} \boldsymbol{\tilde{A}} \boldsymbol{\tilde{D}}^{-\frac{1}{2}} \boldsymbol{H}^{(l)} \boldsymbol{W}^{(l)}\right),
\end{equation}
where $\boldsymbol{\tilde{A}} \in \mathbb{R}^{n \times n}$ is the transformed adjacency matrix $ \boldsymbol{A^{\prime}} $ with self-connections, $ \boldsymbol{\tilde{D}} $ is the degree matrix of $\boldsymbol{\tilde{A}}$ and $ \boldsymbol{W}^{(l)} $ is a learnable weight matrix. The input of GCNs is the node embedding $\boldsymbol{Z} $, and the output is the final feature $\boldsymbol{W} \in \mathbb{R}^{n \times D}$ of label node, where the $D$ represent the dimensionality of image extracted features. Since $ \boldsymbol{A^{\prime}} $ is learned from graph attention transformer layer, our final output can learn the characteristics of the node from a more diverse label node representation space. In general, our label relation branch contains the generated matrix, graph attention transformer layer and GCNs, which is shown in Fig.~\ref{fig1}.

For the image feature branch, let $\boldsymbol{I} $ denote an image, and common CNN networks can be used to extract high-dimensional features of the image. Subsequently, global max-pooling will be adopted to obtain the one-dimensional feature $\boldsymbol{x} \in \mathbb{R}^{D} $ of the image:

\begin{equation}
	\boldsymbol{x}=f_{\mathrm{GMP}}\left(f_{\mathrm{CNN}}\left(\boldsymbol{I} ; \theta_{\mathrm{CNN}}\right)\right),
\end{equation}
where $ \theta_{\mathrm{CNN}} $ denotes the parameters of CNN.

Generally, our entire network include two branches. The image feature branch takes image $\boldsymbol{I} $ as input and output extracted feature $ \boldsymbol{x} $. The label relation branch inputs node embedding $ \boldsymbol{Z} $, and output learned label representation $ \boldsymbol{W} $. Thus, we can get predicted score $ \boldsymbol{\hat{y}} \in \mathbb{R}^{n} $:

\begin{equation}
 \boldsymbol{\hat{y}}=\boldsymbol{W} \boldsymbol{x} .
\end{equation}

Let $ \boldsymbol{y} \in \mathbb{R}^{n} $ denote the ground-truth label space with $ n $ class labels, and $ y^{i}=\{0,1\}$ represents whether the $i$th label appears or not. The optimization goal is to minimize the following loss:

\begin{equation}
	\mathcal{L}=\sum_{i=1}^{n} y^{i} \log \left(\sigma\left(\hat{y}^{i}\right)\right)+\left(1-y^{i}\right) \log \left(1-\sigma\left(\hat{y}^{i}\right)\right),
\end{equation}
where $\sigma(\cdot)$ usually use sigmoid function. 

\section{Experiments}
\label{gen_inst}

In this section, we firstly introduce datasets and compare our method with state-of-the-art models. Then we conduct experiments with different CNN feature extractors to study the node representation learning ability of our proposed method. Subsequently, we visualize the generated correlation matrix by building a relation graph, which vividly shows the difference between our generated matrix and co-occurrence matrix. Then, we study the adaptation ability of our proposed graph attention transformer layer. Finally, the ablation experiments aim to explore the importance of each module.

\begin{table*}\footnotesize
	\setlength\tabcolsep{1.3pt}
	\caption{Comparison with state-of-the-art models in VOC2007}
	\label{voccmp}
	\centering
	\begin{tabular}{ccccccccccccccccccccccc}
		\toprule
		Methods    & aero & bike & bird & boat & bottle & bus & car & cat & chair & cow & table & dog & horse & motor & person & plant & sheep & sofa & train & tv & mAP \\
		\midrule
		CNN-RNN \cite{wang2016cnn} & 96.7 & 83.1 & 94.2 & 92.8 & 61.2 & 82.1 & 89.1 & 94.2 & 64.2 & 83.6 & 70.0 & 92.4 & 91.7 & 84.2 & 93.7 & 59.8 & 93.2 & 75.3 & \bf{99.7} & 78.6 & 84.0 \\
		ResNet-101 \cite{he2016deep} & 99.5 & 97.7 & 97.8 & 96.4 & 65.7 & 91.8 & 96.1 & 97.6 & 74.2 & 80.9 & 85.0 & 98.4 & 96.5 & 95.9 & 98.4 & 70.1 & 88.3 & 80.2 & 98.9 & 89.2 & 89.9 \\
		FeV+LV \cite{yang2016exploit} & 97.9 & 97.0 & 96.6 & 94.6 & 73.6 & 93.9 & 96.5 & 95.5 & 73.7 & 90.3 & 82.8 & 95.4 & 97.7 & 95.9 & 98.6 & 77.6 & 88.7 & 78.0 & 98.3 & 89.0 & 90.6 \\
		HCP \cite{wei2016hcp} & 98.6 & 97.1 & 98.0 & 95.6 & 75.3 & 94.7 & 95.8 & 97.3 & 73.1 & 90.2 & 80.0 & 97.3 & 96.1 & 94.9 & 96.3 & 78.3 & 94.7 & 76.2 & 97.9 & 91.5 & 90.9 \\
		RNN-Attention \cite{wang2017multi} & 98.6 & 97.4 & 96.3 & 96.2 & 75.2 & 92.4 & 96.5 & 97.1 & 76.5 & 92.0 & 87.7 & 96.8 & 97.5 & 93.8 & 98.5 & 81.6 &  93.7 & 82.8 & 98.6 & 89.3 & 91.9 \\
		Atten-Reinforce \cite{chen2017recurrent} & 98.6 & 97.1 & 97.1 & 95.5 & 75.6 & 92.8 & 96.8 & 97.3 & 78.3 & 92.2 & 87.6 & 96.9 & 96.5 & 93.6 & 98.5 & 81.6 & 93.1 & 83.2 & 98.5 & 89.3 & 92.0 \\
		VGG \cite{simonyan2015very} & 99.4 & 97.4 & 98.0 & 97.0 & 77.9 & 92.4 & 96.8 & 97.8 & 80.8 & 93.4 & 87.2 & 98.0 & 97.3 & 95.8 & 98.8 & 79.4 & 95.3 & 82.2 & 99.1 & 91.4 & 92.8 \\
		DLDL \cite{gao2017deep} & 99.3 & 97.6 & 98.3 & 97.0 & 79.0 & 95.7 & 97.0 & 97.9 & 81.8 & 93.3 & 88.2 & 98.1 & 96.9 & 96.5 & 98.4 & 84.8 & 94.9 & 82.7 & 98.5 & 92.8 & 93.4 \\
		ML-GCN \cite{chen2019multi} & 99.5 & 98.5 & 98.6 & 98.1 & 80.8 & 94.6 & 97.2 & 98.2 & 82.3 & 95.7 & 86.4 & 98.2 & 98.4 & 96.7 & 99.0 & 84.7 & 96.7 & 84.3 & 98.9 & 93.7 & 94.0 \\
		MGTN \cite{NguyenAAAI2021} & 99.5 & 98.3 & 98.8 & 97.7 & 86.3 & 97.2 & \bf{98.2} & 99.0 & 83.0 & 96.7 & 90.4 & 97.5 & 98.9 & 97.6 & \bf{99.1} & 87.5 & 97.9 & 84.6 & 98.6 & 95.2 & 95.1   \\
	\midrule
	\bf{GATN} & \bf{99.8} & \bf{98.9} & \bf{99.1} & \bf{98.8} & \bf{89.5} & \bf{97.6} & 97.6 & \bf{99.3} & \bf{87.3} & \bf{98.4} & \bf{90.7} & \bf{99.1} & \bf{99.2} & \bf{98.5} & \bf{99.1} & \bf{88.7} & \bf{98.4} & \bf{90.8} & 99.1 & \bf{96.3} & \bf{96.3}  \\
		\bottomrule
	\end{tabular}
\end{table*}

\begin{table*}
	\caption{Comparison with state-of-the-art models in MS-COCO}
	\label{mscococmp}
	\centering
	\begin{tabular}{cccccccc}
		\toprule
		Methods     & mAP & CP & CR & CF1 & OP & OR & OF1 \\
		\midrule
SRN \cite{zhu2017learning} & 77.1 & 81.6 & 65.4 & 71.2 & 82.7 & 69.9 & 75.8 \\
ResNet-101 \cite{he2016deep} & 77.3 & 80.2 & 66.7 & 72.8 & 83.9 & 70.8 & 76.8 \\
A-GCN \cite{li2019learning} & 83.1 & 84.7 & 72.3 & 78.0 & 85.6 & 75.5 & 80.3 \\
ML-GCN \cite{chen2019multi}  & 83.4 & 83.0 & 73.7 & 78.0 & 82.9 & 76.6 & 79.6 \\
KSSNET \cite{wang2020multi} & 83.7 & 84.6 & 73.2 & 77.2 & 87.8 & 76.2 & 81.5 \\
KGGR \cite{chen2020knowledge} & 84.3 & 85.6 & 72.7 & 78.6 & 87.1 & 75.6 & 80.9\\
C-Tran \cite{lanchantin2020general} & 85.1 & 86.3 & 74.3 & 79.9 & 87.7 & 76.5 & 81.7 \\

MGTN \cite{NguyenAAAI2021} & 87.4 & 86.2 & 79.5 & 82.7 & 87.0 & 81.3 & 84.1 \\
\midrule
\bf{GATN}  & \bf{89.3} & \bf{89.1} & \bf{79.9} & \bf{84.3} & \bf{89.6} & \bf{82.0} & \bf{85.7} \\
		\bottomrule
	\end{tabular}
\end{table*}

\subsection{Datasets and Evaluation Metrics}
\label{sec41}
We evaluate our method on three datasets: PASCAL VOC2007 \cite{everingham2010the}, MS-COCO \cite{lin2014microsoft} and NUS-WIDE \cite{chua2009nus}. VOC2007 has 9,963 images in 20 classes, and each image is assigned with one or multiple labels. We use 5,011 images for training and 4,952 images for testing. MS-COCO is a challenging multi-label classification dataset which consists of 80 classes. In this dataset, the training set includes 82,081 images, and the testing set includes 40,137 images. NUS-WIDE is a much larger dataset with 269,648 images. Following \cite{liu2017semantic, yazici2020orderless}, we remove the unreliable images and labels and finally get 208,347 images of 81 categories. Among them, 150,000 images are used for training and 58,347 images for testing. For a fair comparison, three splits are generated and the average score of each split is calculated in experiments, which follows the split settings of \cite{liu2017semantic}.

\textbf{Evaluation metrics}: We use average precision (AP), mean average precision (mAP), the average overall precision (OP), recall (OR), F1 (OF1) and average per-class precision (CP), recall (CR), F1 (CF1) to measure our models. In general, mean average precision (\textbf{mAP}), the average overall F1 (\textbf{OF1}), and average per-class F1 (\textbf{CF1}) are more important for measuring model's performance.  

\subsection{Comparisons with state-of-the-arts}
\label{sec42}

Our proposed model has two GCN layers, and the output dimensionality is 1,024 and 2,048, respectively. The hidden size in the attention layer is set the same as the number of classes. The threshold $\tau$ in Eq(2) and the $ k $ in Eq(7) are empirically set as 0.2 and 2. We employ ResNeXt-101 32X16d as backbone \cite{xie2017aggregated} for image feature extraction with a semi-weakly supervised pre-trained model on ImageNet \cite{yalniz2019billion}. For a fair comparison with the previous method, we also use GloVe \cite{pennington2014glove} model as our label embeddings. The whole model is trained with SGD with momentum set as 0.9 and learning rate set as 0.03, which has a 0.1 weight decay. The model is trained for 50 epochs with batch size 16. 

\begin{table*}
	\caption{Comparison with state-of-the-art models in NUS-WIDE}
	\label{nuswidecmp}
	\centering
	\begin{tabular}{cccccccc}
		\toprule
		Methods     & mAP & CP & CR & CF1 & OP & OR & OF1 \\
		\midrule
 CNN-RNN \cite{wang2016cnn}  & - & 40.5 & 30.4 & 34.7 & 49.9 & 61.7 & 55.2  \\ 
  Li et al. \cite{li2018attentive} & - & 44.2 & 49.3 & 46.6 & 53.9 & 68.7 & 60.4 \\
  SR CNN-RNN \cite{liu2017semantic} & - & 55.7 & 50.2 & 52.8 & 70.6 & \bf{71.4} & \bf{71.0} \\
 Chen et al. \cite{chen2017order} & - & 59.4 & 50.7 & 54.7 & 69.0 & \bf{71.4} & 70.2 \\
 ML-GCN \cite{chen2019multi} & 54.9 & \bf{65.2} & 43.0 & 51.6 & \bf{75.1} & 62.5 & 68.1   \\
 MGTN \cite{NguyenAAAI2021} & 57.4 & 58.6 & \bf{55.1} & 56.8 & 71.6 & 67.0 & 69.2 \\
\midrule
\bf{GATN}  & \bf{59.8} & 65.0 & 50.7 & \bf{56.9} & 73.4 & 68.1 & 70.7 \\
		\bottomrule
	\end{tabular}
\end{table*}

\begin{table*}
	\caption{Comparison under different backbones}
	\label{backbonecmp}
	\centering
	\begin{tabular}{lllllllll}
				\toprule
		Backbone                                & model  & mAP    & CP    & CR    & CF1   & OP    & OR    & OF1   \\
		\midrule
		\multirow{3}{*}{ResNet101}              & ML-GCN & 83.0 & \bf{85.1} & 72.0 & \bf{78.0} & 85.8 & 75.4 & 80.3  \\
		& MGTN   & \bf{83.1} & 84.0 & \bf{72.8} & \bf{78.0} & 85.4 & \bf{75.8} & 80.3 \\
		& GATN   & 82.9 & 83.9 & 71.8  & 77.4  & \bf{86.3} & 75.4 & \bf{80.4}      \\
		\midrule
		\multirow{3}{*}{ResNeXt-50 32x4d SWSL}   & ML-GCN & 83.4 & 83.0 & 73.7 & 78.0 & 82.9 & 76.6 & 79.6 \\
		& MGTN   & 86.9 & 86.3 & \bf{77.4} & 81.6 & 87.5 & \bf{79.4} & 83.2 \\
		& GATN   & \bf{87.1} & \bf{87.0} & 77.3 & \bf{81.8} & \bf{88.1} & \bf{79.4} & \bf{83.5} \\
		\midrule
		\multirow{3}{*}{ResNeXt-101 32x4d SWSL}  & ML-GCN & 79.4 & 79.8 & 69.3 & 74.2 & 81.2 & 74.4 & 77.7      \\
		& MGTN   & 87.4 & 85.1 & 78.6 & 81.7 & 84.9 & \bf{81.0} & 82.9  \\
		& GATN   & \bf{87.9} & \bf{87.6} & \bf{78.7}  & \bf{82.9}  & \bf{88.0} & 80.9 & \bf{84.3} \\
		\midrule
		\multirow{3}{*}{ResNeXt-101 32x8d SWSL}  & ML-GCN & 80.2 & 80.2 & 70.1 & 74.8 & 82.6 & 74.5 & 78.4  \\
		& MGTN   & 87.0 & 85.8 & 78.4 & 81.9 & 86.7 & 80.6 & 83.5 \\
		& GATN   & \bf{88.0} & \bf{87.6} & \bf{79.0} & \bf{83.1} & \bf{88.8} & \bf{81.9} & \bf{85.2} \\
		\midrule
		\multirow{3}{*}{ResNeXt-101 32x16d SWSL} & ML-GCN & 80.9 & 79.6 & 72.0 & 75.6 & 82.6 & 75.9 & 79.1       \\
		& MGTN   & 87.4 & 86.2 & 79.5 & 82.7 & 87.0    & 81.3 & 84.1 \\
		& GATN   & \bf{89.3}  & \bf{89.1} & \bf{79.9} & \bf{84.3} & \bf{89.6}  & \bf{82.0} & \bf{85.7} \\

		\bottomrule
	\end{tabular}
\end{table*}

For VOC2007 dataset, we compare our model with following state-of-the-art models: CNN-RNN \cite{wang2016cnn}, ResNet-101 \cite{he2016deep}, 	FeV+LV \cite{yang2016exploit}, HCP \cite{wei2016hcp}, RNN-Attention \cite{wang2017multi}, Atten-Reinforce \cite{chen2017recurrent}, VGG \cite{simonyan2015very}, DLDL \cite{gao2017deep}, ML-GCN \cite{chen2019multi} and MGTN \cite{NguyenAAAI2021}. As shown in Table~\ref{voccmp}, our proposed model reach optimal performance. GATN outperform ResNet-101 6.4\%, VGG 3.5\% in mAP, while ResNet-101 and VGG are considered mainstream deep CNN models. For RNN-based methods, GATN shows 12.3\% promotion of mAP over CNN-RNN, and it also holds a 4.4\% and 4.3\% higher mAP over two methods RNN-Attention and Atten-Reinforce, which also include attention mechanisms. In comparison, GNN-based methods have a better performance than RNN-based methods. Two GNN-based methods ML-GCN and MGTN both have more than 94\% mAP. Compared with them, GATN has a 2.3\% and 1.2\% improvements in mAP over ML-GCN and MGTN. Besides, specific to each category, GATN achieve state-of-the-art performance on 18 categories in AP. Except for \emph{car} and \emph{train}, our model all reach the current best AP. Especially for \emph{bottle}, \emph{chair} and \emph{sofa}, our method can exceed the previous state-of-the-art result by more than 3\% AP.

In Table~\ref{mscococmp}, we compare our model with following state-of-the-art models in MS-COCO dataset: SRN \cite{zhu2017learning}, ResNet-101 \cite{he2016deep}, KGGR \cite{chen2020knowledge}, ML-GCN \cite{chen2019multi}, A-GCN \cite{li2019learning}, KSSNET \cite{wang2020multi}, C-Tran \cite{lanchantin2020general} and MGTN \cite{NguyenAAAI2021}. We can see that GATN can outperform other methods in almost all metrics. It has a 12.2\% and 12.0\% higher mAP, 13.1\% and 11.5\% higher CF1, 9.9\% and 8.9\% higher OF1 than SRN and ResNet-101, respectively. Compared with GNN-based methods, GATN can also exceed ML-GCN 5.9\%, 6.3\%, 6.1\% in mAP, CF1, OF1, and show an improvement of 1.9\%, 1.6\%, 1.6\% in mAP, CF1, OF1 over MGTN, while the latter is regarded as the most powerful model currently. Moreover, C-Tran is a transformer-based method, and GATN also shows a significant advantage over it, which shows that applying the self-attention mechanism to graphs is a more effective way. Lastly, GATN can achieve state-of-the-art performance in all seven metrics.

As shown in Table~\ref{nuswidecmp}, we compare GATN with the competitive models including CNN-RNN \cite{wang2016cnn}, Li et al. \cite{li2018attentive}, SR CNN-RNN \cite{liu2017semantic}, Chen et al. \cite{chen2017order}, ML-GCN \cite{chen2019multi} and MGTN \cite{NguyenAAAI2021}. Owing to a large amount of data in NUS-WIDE and the division method is not unique, different models have their own merits on different indicators. In general, GATN holds the best performance in mAP, CF1, and the second-best performance in CP, CR, OP, and OF1. 

\begin{figure*}
	\centering
	\includegraphics[width=\textwidth]{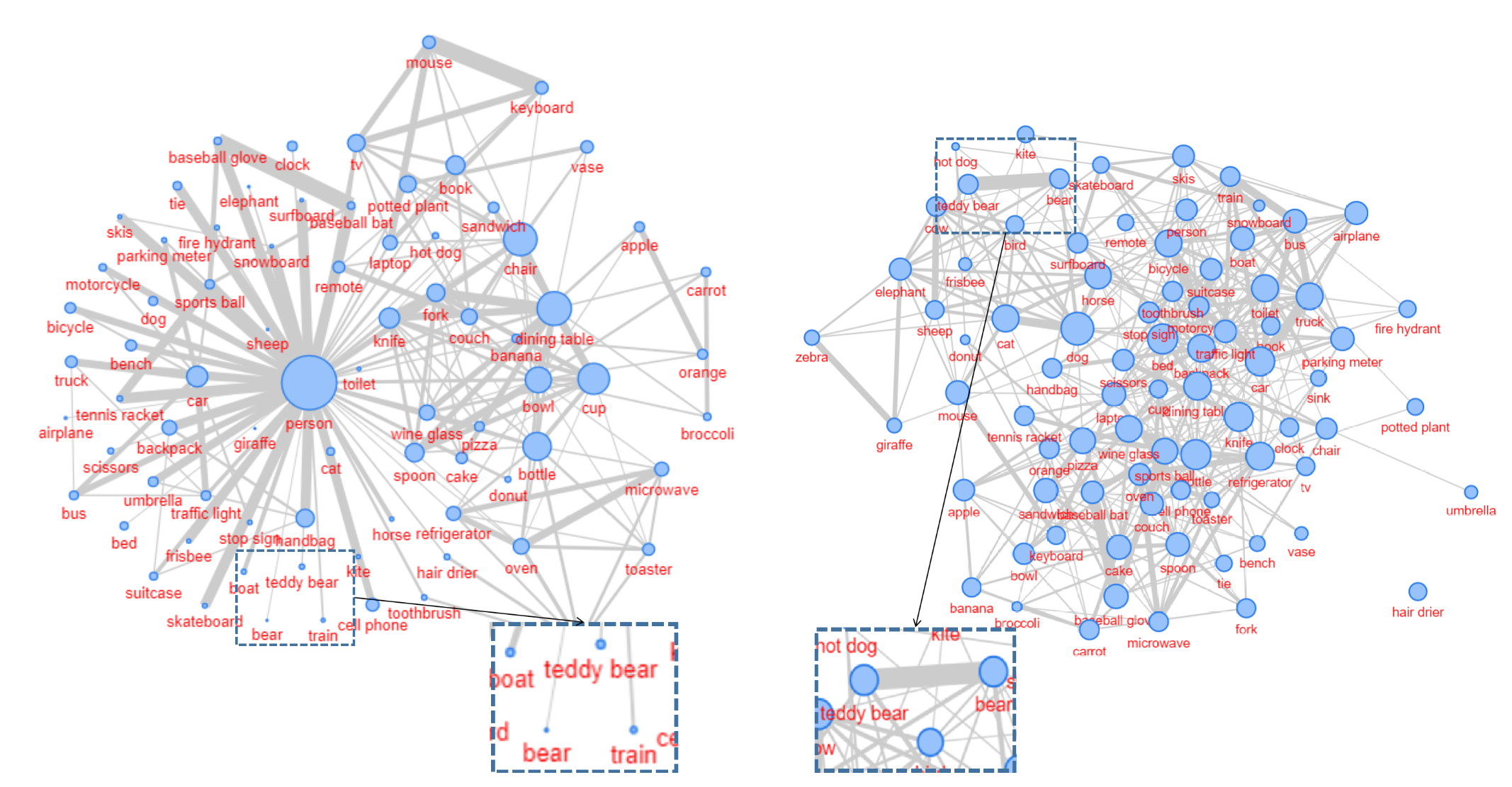}
	\caption{Visualization of the adjacency matrix on the MS-COCO dataset. Each blue circle is a label node whose size is determined by the sum of its relation values. The gray line represents the relationship, and the thicker line means the stronger relationship. The left picture is the co-occurrence matrix and the right picture is our generated correlation matrix. Specifically, we zoomed in \emph{bear} and \emph{teddy bear} for a comparison.}
	\label{fig3}
\end{figure*}

\subsection{Model Representation Ability Experiments}

GNN-based methods rely on the extraction of image features and label correlation features. To further explore the importance of label correlation features and the label correlation representation ability of different GNN models, we compare our model with two GNN-based models (ML-GCN and MGTN) under different backbones. 

As shown in Table~\ref{backbonecmp}, for ResNet-101, these three models have similar performance, including mAP around 83.0, CF1 around 78.0 and OF1 around 80.3, which shows that the performance of the model at this time is limited by the image extractor. With the increase of the representation ability of image feature backbone models, ML-GCN has a slight improvement with ResNeXt-50 32x4d SWSL of 0.4\% mAP. At this time, MGTN and GATN both have a greater improvement. MGTN has 3.8\%, 3.6\% and 2.9\% improvements on mAP, OF1, and CF1, while GATN gains 4.2\%, 4.4\% and 3.1\% in these three metrics. For this backbone, ML-GCN is limited by its label correlation feature learning branch, which also illustrates that the core of the multi-label task is to learn more powerful relational representations. On the contrary, MGTN and GATN both get better results with the increase of image feature expression ability, which can be attributed to their powerful label correlation features.

For ResNeXt-101 models, ML-GCN has some decline in all metrics, so we will not discuss it later. On the other hand, MGTN and GATN both get better results by using ResNeXt-101. However, GATN continues to improve its performance as the backbone model's ability become more powerful, it has a 0.8\% increase of mAP, 1.1\% increase of CF1, 0.8\% increase of OF1 with ResNeXt-101 32x4d SWSL compared with ResNeXt-50, and it also obtains 2.2\% mAP increment, 2.5\% CF1 increment and 2.2 \% OF1 increment with ResNeXt-101 32x16d SWSL compared with ResNeXt-50. Different from GATN, MGTN has little difference in performance on different ResNeXt-101, it only has a 1.0\% and 1.2\% rise in CF1 and OF1 when comparing its result on ResNeXt-101 32x4d SWSL and ResNeXt-101 32x16d SWSL. These results demonstrate that GATN has a stronger representation ability in label correlation learning compared to MGTN.

\subsection{Visualization of the Generated Correlation Matrix}

To further demonstrate the effect of our generated correlation matrix, we conduct a visualization experiment for the MS-COCO dataset in this section, including 80 categories.

As shown in Fig.~\ref{fig3}, we compare our generated correlation matrix with the co-occurrence matrix. Firstly, we set node size as the sum of all its relations. Thus, the larger the node, the more the number of its relationship with other categories. Subsequently, the thickness of the edge is defined as the strength of the relationship, and the thick edge indicates a strong relationship. To avoid visual disturbance caused by too many edges, we filter the edges with a threshold value of 0.25.

Comparing the two topology graphs, we can find \emph{person} occupies a dominant position due to its extremely high frequency in the training set. On the contrary, our generated correlation matrix has a more balanced relationship strength. Specifically, we can observe that \emph{teddy bear} and \emph{bear} have a strong relation in our generated matrix since they have a similar appearance. However, the two rarely appear simultaneously in natural images, which leads to the co-occurrence matrix being likely to ignore this relation. In general, the correlation matrix we generated can capture richer semantic information, and the graph attention transformer layer can select the information based on this semantic information to adapt to different feature extraction models.

\subsection{Adaptation Ability Study}

To study the adaptation ability of our proposed graph attention transformer layer, we output the adjacency matrix before and after our graph attention transformer layer for comparison.

\begin{figure}
	\centering
	\includegraphics[width=\columnwidth]{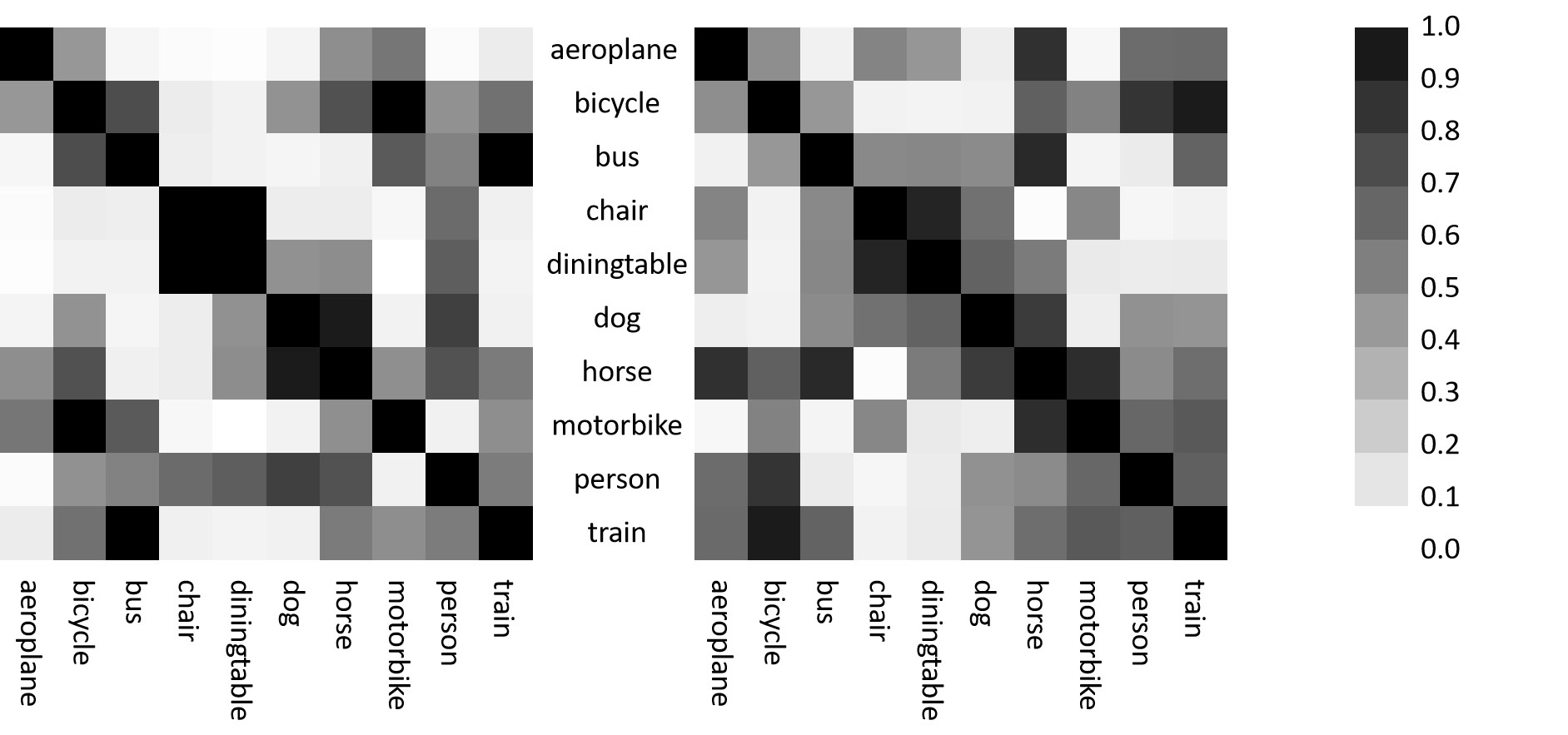}
	\caption{The left picture is the initialized correlation matrix generated by the distance-based method, and the right is the transformed matrix after the graph attention transformer layer. The darker the color, the stronger the relationship between the two categories.}
	\label{fig31}
\end{figure}

We randomly select 10 categories in the VOC2007 dataset and show results in Fig.~\ref{fig31}. We can find the relation between \emph{dog} and \emph{chair} become stronger, which can be attributed to the VOC2007 dataset containing a large number of photos of dogs at home. Samely, we can find in the distance-based matrix, the relation between \emph{person} and \emph{aeroplane} is lower than that between \emph{person} and \emph{bus}, which is reasonable since people usually take buses more often than airplanes. Due to the photos of airplanes and people in the VOC2007 dataset being relatively high, the relationship between \emph{airplane} and \emph{person} in the transformed matrix is becoming stronger. Moreover, \emph{train} and \emph{bus} are deeply related in semantic information because they are both vehicles. In this dataset, the two often appear in isolation, which makes the relationship between the \emph{train} and \emph{bus} in the transformed matrix weaker.

\subsection{Ablation Studies}

To verify the effectiveness of each module of our proposed method, we conduct ablation studies. Since our method is based on ML-GCN, we name the original ML-GCN as ML-GCN (CO-OCC), the version that only uses our generated correlation matrix as ML-GCN (CORR), the version that only uses graph attention layer as GATN (CO-OCC), and our final version is named GATN (CORR). All experiments are conducted in the MS-COCO dataset.

As shown in Table~\ref{abla}, both ML-GCN (CORR) and GATN (CO-OCC) show advantages over ML-GCN (CO-OCC). ML-GCN (CORR) has a 0.9\%, 1.1\% and 1.6\% improvements on mAP, CF1 and OF1, while GATN (CO-OCC) has 5.2\%, 5.2\% and 3.6\% improvements on mAP, CF1 and OF1. Generally, our proposed graph attention transformer layer is more helpful for performance, which is reasonable since the generated matrix indeed changes the initial graph adjacency matrix without changing the adaptation ability of the model. Moreover, we can find the GATN (CORR) also have a 0.7\%, 1.1\% and 2.5\% in mAP, CF1 and OF1 compared to GATN (CO-OCC), further explaining that the generated correlation matrix is also helpful to the result.

\begin{table}\small
\renewcommand\tabcolsep{2.0pt}
	\caption{Ablation studies}
	\label{abla}
	\centering
	\begin{tabular}{cccccccc}
		\toprule
		Methods     & mAP & CP & CR & CF1 & OP & OR & OF1 \\
		\midrule
ML-GCN (CO-OCC)  & 83.4 & 83.0 & 73.7 & 78.0 & 82.9 & 76.6 & 79.6  \\
ML-GCN (CORR) & 84.3 & 86.9 & 72.6 & 79.1 & 87.4 & 75.9 & 81.2 \\
GATN (CO-OCC) & 88.6 & 85.6 & 80.9 & 83.2 & 84.0 & 82.4 & 83.2 \\
GATN (CORR) & 89.3 & 89.1 & 79.9 & 84.3 & 89.6 & 82.0 & 85.7 \\
		\bottomrule
	\end{tabular}
\end{table}

\section{Conclusion}

In this paper, we propose Graph Attention Transformer Networks to achieve multi-label image classification. Our proposed method generates a correlation matrix through a distance-based algorithm and exploits a novel structure including several self-attention branches to mine the multi-hop paths in the correlation matrix. By transforming the correlation matrix into multiple sub-graphs, we can improve the expressive ability of the label relation, thereby improving the accuracy of multi-label classification tasks. A large number of experiments on multiple datasets prove the effectiveness of our method. We believe that our method has powerful abilities. It only needs to input the node representation in the network, and then it can extract the different transformations of relationships among nodes. In the future, we will continue to explore the performance of this network in other tasks, such as node classification and relationship extraction.




\begin{thebibliography}{50}


\ifx \showCODEN    \undefined \def \showCODEN     #1{\unskip}     \fi
\ifx \showDOI      \undefined \def \showDOI       #1{#1}\fi
\ifx \showISBNx    \undefined \def \showISBNx     #1{\unskip}     \fi
\ifx \showISBNxiii \undefined \def \showISBNxiii  #1{\unskip}     \fi
\ifx \showISSN     \undefined \def \showISSN      #1{\unskip}     \fi
\ifx \showLCCN     \undefined \def \showLCCN      #1{\unskip}     \fi
\ifx \shownote     \undefined \def \shownote      #1{#1}          \fi
\ifx \showarticletitle \undefined \def \showarticletitle #1{#1}   \fi
\ifx \showURL      \undefined \def \showURL       {\relax}        \fi
\providecommand\bibfield[2]{#2}
\providecommand\bibinfo[2]{#2}
\providecommand\natexlab[1]{#1}
\providecommand\showeprint[2][]{arXiv:#2}

\bibitem[{Bahdanau} et~al\mbox{.}(2015)]%
        {bahdanau2015neural}
\bibfield{author}{\bibinfo{person}{Dzmitry {Bahdanau}},
  \bibinfo{person}{Kyunghyun {Cho}}, {and} \bibinfo{person}{Yoshua {Bengio}}.}
  \bibinfo{year}{2015}\natexlab{}.
\newblock \showarticletitle{Neural Machine Translation by Jointly Learning to
  Align and Translate}. In \bibinfo{booktitle}{\emph{ICLR 2015 : International
  Conference on Learning Representations 2015}}.
\newblock


\bibitem[{Bruna} et~al\mbox{.}(2014)]%
        {bruna2014spectral}
\bibfield{author}{\bibinfo{person}{Joan {Bruna}}, \bibinfo{person}{Wojciech
  {Zaremba}}, \bibinfo{person}{Arthur {Szlam}}, {and} \bibinfo{person}{Yann
  {LeCun}}.} \bibinfo{year}{2014}\natexlab{}.
\newblock \showarticletitle{Spectral Networks and Locally Connected Networks on
  Graphs}. In \bibinfo{booktitle}{\emph{ICLR 2014 : International Conference on
  Learning Representations (ICLR) 2014}}.
\newblock


\bibitem[{Chen} et~al\mbox{.}(2020b)]%
        {chen2020generative}
\bibfield{author}{\bibinfo{person}{Mark {Chen}}, \bibinfo{person}{Alec
  {Radford}}, \bibinfo{person}{Rewon {Child}}, \bibinfo{person}{Jeffrey~K
  {Wu}}, \bibinfo{person}{Heewoo {Jun}}, \bibinfo{person}{David {Luan}}, {and}
  \bibinfo{person}{Ilya {Sutskever}}.} \bibinfo{year}{2020}\natexlab{b}.
\newblock \showarticletitle{Generative Pretraining From Pixels}. In
  \bibinfo{booktitle}{\emph{ICML 2020: 37th International Conference on Machine
  Learning}}, Vol.~\bibinfo{volume}{1}. \bibinfo{pages}{1691--1703}.
\newblock


\bibitem[{Chen} et~al\mbox{.}(2020c)]%
        {chen2020label}
\bibfield{author}{\bibinfo{person}{Shikai {Chen}}, \bibinfo{person}{Jianfeng
  {Wang}}, \bibinfo{person}{Yuedong {Chen}}, \bibinfo{person}{Zhongchao {Shi}},
  \bibinfo{person}{Xin {Geng}}, {and} \bibinfo{person}{Yong {Rui}}.}
  \bibinfo{year}{2020}\natexlab{c}.
\newblock \showarticletitle{Label Distribution Learning on Auxiliary Label
  Space Graphs for Facial Expression Recognition}. In
  \bibinfo{booktitle}{\emph{2020 IEEE/CVF Conference on Computer Vision and
  Pattern Recognition (CVPR)}}. \bibinfo{pages}{13984--13993}.
\newblock


\bibitem[{Chen} et~al\mbox{.}(2017a)]%
        {chen2017order}
\bibfield{author}{\bibinfo{person}{Shang-Fu {Chen}}, \bibinfo{person}{Yi-Chen
  {Chen}}, \bibinfo{person}{Chih-Kuan {Yeh}}, {and}
  \bibinfo{person}{Yu-Chiang~Frank {Wang}}.} \bibinfo{year}{2017}\natexlab{a}.
\newblock \showarticletitle{Order-Free RNN with Visual Attention for
  Multi-Label Classification}. In \bibinfo{booktitle}{\emph{AAAI}}.
  \bibinfo{pages}{6714--6721}.
\newblock


\bibitem[{Chen} et~al\mbox{.}(2020a)]%
        {chen2020knowledge}
\bibfield{author}{\bibinfo{person}{Tianshui {Chen}}, \bibinfo{person}{Liang
  {Lin}}, \bibinfo{person}{Xiaolu {Hui}}, \bibinfo{person}{Riquan {Chen}},
  {and} \bibinfo{person}{Hefeng {Wu}}.} \bibinfo{year}{2020}\natexlab{a}.
\newblock \showarticletitle{Knowledge-Guided Multi-Label Few-Shot Learning for
  General Image Recognition.}
\newblock \bibinfo{journal}{\emph{IEEE Transactions on Pattern Analysis and
  Machine Intelligence}} (\bibinfo{year}{2020}), \bibinfo{pages}{1--1}.
\newblock


\bibitem[{Chen} et~al\mbox{.}(2017b)]%
        {chen2017recurrent}
\bibfield{author}{\bibinfo{person}{Tianshui {Chen}}, \bibinfo{person}{Zhouxia
  {Wang}}, \bibinfo{person}{Guanbin {Li}}, {and} \bibinfo{person}{Liang
  {Lin}}.} \bibinfo{year}{2017}\natexlab{b}.
\newblock \showarticletitle{Recurrent Attentional Reinforcement Learning for
  Multi-label Image Recognition}. In \bibinfo{booktitle}{\emph{AAAI}}.
  \bibinfo{pages}{6730--6737}.
\newblock


\bibitem[{Chen} et~al\mbox{.}(2019)]%
        {chen2019multi}
\bibfield{author}{\bibinfo{person}{Zhao-Min {Chen}}, \bibinfo{person}{Xiu-Shen
  {Wei}}, \bibinfo{person}{Peng {Wang}}, {and} \bibinfo{person}{Yanwen {Guo}}.}
  \bibinfo{year}{2019}\natexlab{}.
\newblock \showarticletitle{Multi-Label Recognition With Graph Convolutional
  Networks}. In \bibinfo{booktitle}{\emph{2019 IEEE/CVF Conference on Computer
  Vision and Pattern Recognition (CVPR)}}. \bibinfo{pages}{5177--5186}.
\newblock


\bibitem[{Chua} et~al\mbox{.}(2009)]%
        {chua2009nus}
\bibfield{author}{\bibinfo{person}{Tat-Seng {Chua}}, \bibinfo{person}{Jinhui
  {Tang}}, \bibinfo{person}{Richang {Hong}}, \bibinfo{person}{Haojie {Li}},
  \bibinfo{person}{Zhiping {Luo}}, {and} \bibinfo{person}{Yantao {Zheng}}.}
  \bibinfo{year}{2009}\natexlab{}.
\newblock \showarticletitle{NUS-WIDE: a real-world web image database from
  National University of Singapore}. In \bibinfo{booktitle}{\emph{Proceedings
  of the ACM International Conference on Image and Video Retrieval}}.
  \bibinfo{pages}{48}.
\newblock


\bibitem[{Devlin} et~al\mbox{.}(2018)]%
        {devlin2018bert}
\bibfield{author}{\bibinfo{person}{Jacob {Devlin}}, \bibinfo{person}{Ming-Wei
  {Chang}}, \bibinfo{person}{Kenton {Lee}}, {and} \bibinfo{person}{Kristina~N.
  {Toutanova}}.} \bibinfo{year}{2018}\natexlab{}.
\newblock \showarticletitle{BERT: Pre-training of Deep Bidirectional
  Transformers for Language Understanding}. In
  \bibinfo{booktitle}{\emph{Proceedings of the 2019 Conference of the North
  American Chapter of the Association for Computational Linguistics: Human
  Language Technologies, Volume 1 (Long and Short Papers)}}.
  \bibinfo{pages}{4171--4186}.
\newblock


\bibitem[{Dosovitskiy} et~al\mbox{.}(2021)]%
        {dosovitskiy2021an}
\bibfield{author}{\bibinfo{person}{Alexey {Dosovitskiy}},
  \bibinfo{person}{Lucas {Beyer}}, \bibinfo{person}{Alexander {Kolesnikov}},
  \bibinfo{person}{Dirk {Weissenborn}}, \bibinfo{person}{Xiaohua {Zhai}},
  \bibinfo{person}{Thomas {Unterthiner}}, \bibinfo{person}{Mostafa {Dehghani}},
  \bibinfo{person}{Matthias {Minderer}}, \bibinfo{person}{Georg {Heigold}},
  \bibinfo{person}{Sylvain {Gelly}}, \bibinfo{person}{Jakob {Uszkoreit}}, {and}
  \bibinfo{person}{Neil {Houlsby}}.} \bibinfo{year}{2021}\natexlab{}.
\newblock \showarticletitle{An Image is Worth 16x16 Words: Transformers for
  Image Recognition at Scale}. In \bibinfo{booktitle}{\emph{ICLR 2021: The
  Ninth International Conference on Learning Representations}}.
\newblock


\bibitem[{Everingham} et~al\mbox{.}(2010)]%
        {everingham2010the}
\bibfield{author}{\bibinfo{person}{Mark {Everingham}}, \bibinfo{person}{Luc
  {Gool}}, \bibinfo{person}{Christopher~K. {Williams}}, \bibinfo{person}{John
  {Winn}}, {and} \bibinfo{person}{Andrew {Zisserman}}.}
  \bibinfo{year}{2010}\natexlab{}.
\newblock \showarticletitle{The Pascal Visual Object Classes (VOC) Challenge}.
\newblock \bibinfo{journal}{\emph{International Journal of Computer Vision}}
  \bibinfo{volume}{88}, \bibinfo{number}{2} (\bibinfo{year}{2010}),
  \bibinfo{pages}{303--338}.
\newblock


\bibitem[{Gao} et~al\mbox{.}(2017)]%
        {gao2017deep}
\bibfield{author}{\bibinfo{person}{Bin-Bin {Gao}}, \bibinfo{person}{Chao
  {Xing}}, \bibinfo{person}{Chen-Wei {Xie}}, \bibinfo{person}{Jianxin {Wu}},
  {and} \bibinfo{person}{Xin {Geng}}.} \bibinfo{year}{2017}\natexlab{}.
\newblock \showarticletitle{Deep Label Distribution Learning With Label
  Ambiguity}.
\newblock \bibinfo{journal}{\emph{IEEE Transactions on Image Processing}}
  \bibinfo{volume}{26}, \bibinfo{number}{6} (\bibinfo{year}{2017}),
  \bibinfo{pages}{2825--2838}.
\newblock


\bibitem[{Ge} et~al\mbox{.}(2018)]%
        {ge2018chest}
\bibfield{author}{\bibinfo{person}{Zongyuan {Ge}}, \bibinfo{person}{Dwarikanath
  {Mahapatra}}, \bibinfo{person}{Suman {Sedai}}, \bibinfo{person}{Rahil
  {Garnavi}}, {and} \bibinfo{person}{Rajib {Chakravorty}}.}
  \bibinfo{year}{2018}\natexlab{}.
\newblock \showarticletitle{Chest X-rays Classification: A Multi-Label and
  Fine-Grained Problem}.
\newblock \bibinfo{journal}{\emph{arXiv preprint arXiv:1807.07247}}
  (\bibinfo{year}{2018}).
\newblock


\bibitem[{Grover} and {Leskovec}(2016)]%
        {grover2016node2vec}
\bibfield{author}{\bibinfo{person}{Aditya {Grover}} {and} \bibinfo{person}{Jure
  {Leskovec}}.} \bibinfo{year}{2016}\natexlab{}.
\newblock \showarticletitle{node2vec: Scalable Feature Learning for Networks}.
  In \bibinfo{booktitle}{\emph{Proceedings of the 22nd ACM SIGKDD International
  Conference on Knowledge Discovery and Data Mining}},
  Vol.~\bibinfo{volume}{2016}. \bibinfo{pages}{855--864}.
\newblock


\bibitem[{He} et~al\mbox{.}(2016)]%
        {he2016deep}
\bibfield{author}{\bibinfo{person}{Kaiming {He}}, \bibinfo{person}{Xiangyu
  {Zhang}}, \bibinfo{person}{Shaoqing {Ren}}, {and} \bibinfo{person}{Jian
  {Sun}}.} \bibinfo{year}{2016}\natexlab{}.
\newblock \showarticletitle{Deep Residual Learning for Image Recognition}. In
  \bibinfo{booktitle}{\emph{2016 IEEE Conference on Computer Vision and Pattern
  Recognition (CVPR)}}. \bibinfo{pages}{770--778}.
\newblock


\bibitem[{Kipf} and {Welling}(2016)]%
        {kipf2016semi}
\bibfield{author}{\bibinfo{person}{Thomas~N. {Kipf}} {and} \bibinfo{person}{Max
  {Welling}}.} \bibinfo{year}{2016}\natexlab{}.
\newblock \showarticletitle{Semi-Supervised Classification with Graph
  Convolutional Networks}. In \bibinfo{booktitle}{\emph{ICLR (Poster)}}.
\newblock


\bibitem[{Lanchantin} et~al\mbox{.}(2020)]%
        {lanchantin2020general}
\bibfield{author}{\bibinfo{person}{Jack {Lanchantin}}, \bibinfo{person}{Tianlu
  {Wang}}, \bibinfo{person}{Vicente {Ordonez}}, {and} \bibinfo{person}{Yanjun
  {Qi}}.} \bibinfo{year}{2020}\natexlab{}.
\newblock \showarticletitle{General Multi-label Image Classification with
  Transformers.}
\newblock \bibinfo{journal}{\emph{arXiv preprint arXiv:2011.14027}}
  (\bibinfo{year}{2020}).
\newblock


\bibitem[{Lee} et~al\mbox{.}(2018)]%
        {lee2018multi}
\bibfield{author}{\bibinfo{person}{Chung-Wei {Lee}}, \bibinfo{person}{Wei
  {Fang}}, \bibinfo{person}{Chih-Kuan {Yeh}}, {and}
  \bibinfo{person}{Yu-Chiang~Frank {Wang}}.} \bibinfo{year}{2018}\natexlab{}.
\newblock \showarticletitle{Multi-label Zero-Shot Learning with Structured
  Knowledge Graphs}. In \bibinfo{booktitle}{\emph{2018 IEEE/CVF Conference on
  Computer Vision and Pattern Recognition}}. \bibinfo{pages}{1576--1585}.
\newblock


\bibitem[{Li} et~al\mbox{.}(2018)]%
        {li2018attentive}
\bibfield{author}{\bibinfo{person}{Liang {Li}}, \bibinfo{person}{Shuhui
  {Wang}}, \bibinfo{person}{Shuqiang {Jiang}}, {and} \bibinfo{person}{Qingming
  {Huang}}.} \bibinfo{year}{2018}\natexlab{}.
\newblock \showarticletitle{Attentive Recurrent Neural Network for
  Weak-supervised Multi-label Image Classification}. In
  \bibinfo{booktitle}{\emph{Proceedings of the 26th ACM international
  conference on Multimedia}}. \bibinfo{pages}{1092--1100}.
\newblock


\bibitem[{Li} et~al\mbox{.}(2019)]%
        {li2019learning}
\bibfield{author}{\bibinfo{person}{Qing {Li}}, \bibinfo{person}{Xiaojiang
  {Peng}}, \bibinfo{person}{Yu {Qiao}}, {and} \bibinfo{person}{Qiang {Peng}}.}
  \bibinfo{year}{2019}\natexlab{}.
\newblock \showarticletitle{Learning Category Correlations for Multi-label
  Image Recognition with Graph Networks.}
\newblock \bibinfo{journal}{\emph{arXiv preprint arXiv:1909.13005}}
  (\bibinfo{year}{2019}).
\newblock


\bibitem[{Li} et~al\mbox{.}(2016)]%
        {li2016human}
\bibfield{author}{\bibinfo{person}{Yining {Li}}, \bibinfo{person}{Chen
  {Huang}}, \bibinfo{person}{Chen~Change {Loy}}, {and} \bibinfo{person}{Xiaoou
  {Tang}}.} \bibinfo{year}{2016}\natexlab{}.
\newblock \showarticletitle{Human Attribute Recognition by Deep Hierarchical
  Contexts}. In \bibinfo{booktitle}{\emph{European Conference on Computer
  Vision}}. \bibinfo{pages}{684--700}.
\newblock


\bibitem[{Lin} et~al\mbox{.}(2014)]%
        {lin2014microsoft}
\bibfield{author}{\bibinfo{person}{Tsung-Yi {Lin}}, \bibinfo{person}{Michael
  {Maire}}, \bibinfo{person}{Serge~J. {Belongie}}, \bibinfo{person}{James
  {Hays}}, \bibinfo{person}{Pietro {Perona}}, \bibinfo{person}{Deva {Ramanan}},
  \bibinfo{person}{Piotr {Dollár}}, {and} \bibinfo{person}{C.~Lawrence
  {Zitnick}}.} \bibinfo{year}{2014}\natexlab{}.
\newblock \showarticletitle{Microsoft COCO: Common Objects in Context}. In
  \bibinfo{booktitle}{\emph{European Conference on Computer Vision}}.
  \bibinfo{pages}{740--755}.
\newblock


\bibitem[{Liu} et~al\mbox{.}(2017)]%
        {liu2017semantic}
\bibfield{author}{\bibinfo{person}{Feng {Liu}}, \bibinfo{person}{Tao {Xiang}},
  \bibinfo{person}{Timothy~M. {Hospedales}}, \bibinfo{person}{Wankou {Yang}},
  {and} \bibinfo{person}{Changyin {Sun}}.} \bibinfo{year}{2017}\natexlab{}.
\newblock \showarticletitle{Semantic Regularisation for Recurrent Image
  Annotation}. In \bibinfo{booktitle}{\emph{2017 IEEE Conference on Computer
  Vision and Pattern Recognition (CVPR)}}. \bibinfo{pages}{4160--4168}.
\newblock


\bibitem[{Liu} and {Tsang}(2015)]%
        {liu2015on}
\bibfield{author}{\bibinfo{person}{Weiwei {Liu}} {and} \bibinfo{person}{Ivor~W.
  {Tsang}}.} \bibinfo{year}{2015}\natexlab{}.
\newblock \showarticletitle{On the optimality of classifier chain for
  multi-label classification}. In \bibinfo{booktitle}{\emph{NIPS'15 Proceedings
  of the 28th International Conference on Neural Information Processing Systems
  - Volume 1}}, Vol.~\bibinfo{volume}{28}. \bibinfo{pages}{712--720}.
\newblock


\bibitem[Liu et~al\mbox{.}(2021)]%
        {Liu2021survey}
\bibfield{author}{\bibinfo{person}{Yang Liu}, \bibinfo{person}{Yao Zhang},
  \bibinfo{person}{Yixin Wang}, \bibinfo{person}{Feng Hou},
  \bibinfo{person}{Jin Yuan}, \bibinfo{person}{Jiang Tian},
  \bibinfo{person}{Yang Zhang}, \bibinfo{person}{Zhongchao Shi},
  \bibinfo{person}{Jianping Fan}, {and} \bibinfo{person}{Zhiqiang He}.}
  \bibinfo{year}{2021}\natexlab{}.
\newblock \showarticletitle{A survey of visual transformers}.
\newblock \bibinfo{journal}{\emph{arXiv preprint arXiv:2111.06091}}
  (\bibinfo{year}{2021}).
\newblock


\bibitem[{Nam} et~al\mbox{.}(2019)]%
        {nam2019learning}
\bibfield{author}{\bibinfo{person}{Jinseok {Nam}}, \bibinfo{person}{Young-Bum
  {Kim}}, \bibinfo{person}{Eneldo~Loza {Mencia}}, \bibinfo{person}{Sunghyun
  {Park}}, \bibinfo{person}{Ruhi {Sarikaya}}, {and} \bibinfo{person}{Johannes
  {Fürnkranz}}.} \bibinfo{year}{2019}\natexlab{}.
\newblock \showarticletitle{Learning Context-dependent Label Permutations for
  Multi-label Classification}. In \bibinfo{booktitle}{\emph{International
  Conference on Machine Learning}}. \bibinfo{pages}{4733--4742}.
\newblock


\bibitem[{Nath} et~al\mbox{.}(2019)]%
        {nath2019single}
\bibfield{author}{\bibinfo{person}{Nipun~D. {Nath}}, \bibinfo{person}{Theodora
  {Chaspari}}, {and} \bibinfo{person}{Amir~H. {Behzadan}}.}
  \bibinfo{year}{2019}\natexlab{}.
\newblock \showarticletitle{Single- and multi-label classification of
  construction objects using deep transfer learning methods}.
\newblock \bibinfo{journal}{\emph{Journal of Information Technology in
  Construction}} \bibinfo{volume}{24}, \bibinfo{number}{28}
  (\bibinfo{year}{2019}), \bibinfo{pages}{511--526}.
\newblock


\bibitem[Nguyen et~al\mbox{.}(2021)]%
        {NguyenAAAI2021}
\bibfield{author}{\bibinfo{person}{Hoang~D. Nguyen}, \bibinfo{person}{Xuan-Son
  Vu}, {and} \bibinfo{person}{Duc-Trong Le}.} \bibinfo{year}{2021}\natexlab{}.
\newblock \showarticletitle{Modular Graph Transformer Networks for Multi-Label
  Image Classification}. In \bibinfo{booktitle}{\emph{Proceedings of the AAAI
  Conference on Artificial Intelligence}} \emph{(\bibinfo{series}{AAAI '21})}.
  \bibinfo{publisher}{AAAI}.
\newblock


\bibitem[{Pennington} et~al\mbox{.}(2014)]%
        {pennington2014glove}
\bibfield{author}{\bibinfo{person}{Jeffrey {Pennington}},
  \bibinfo{person}{Richard {Socher}}, {and} \bibinfo{person}{Christopher
  {Manning}}.} \bibinfo{year}{2014}\natexlab{}.
\newblock \showarticletitle{Glove: Global Vectors for Word Representation}. In
  \bibinfo{booktitle}{\emph{Proceedings of the 2014 Conference on Empirical
  Methods in Natural Language Processing (EMNLP)}}.
  \bibinfo{pages}{1532--1543}.
\newblock


\bibitem[{Perozzi} et~al\mbox{.}(2014)]%
        {perozzi2014deepwalk}
\bibfield{author}{\bibinfo{person}{Bryan {Perozzi}}, \bibinfo{person}{Rami
  {Al-Rfou}}, {and} \bibinfo{person}{Steven {Skiena}}.}
  \bibinfo{year}{2014}\natexlab{}.
\newblock \showarticletitle{DeepWalk: online learning of social
  representations}. In \bibinfo{booktitle}{\emph{Proceedings of the 20th ACM
  SIGKDD international conference on Knowledge discovery and data mining}}.
  \bibinfo{pages}{701--710}.
\newblock


\bibitem[{Quevedo} et~al\mbox{.}(2012)]%
        {quevedo2012multilabel}
\bibfield{author}{\bibinfo{person}{José~RamóN {Quevedo}},
  \bibinfo{person}{Oscar {Luaces}}, {and} \bibinfo{person}{Antonio
  {Bahamonde}}.} \bibinfo{year}{2012}\natexlab{}.
\newblock \showarticletitle{Multilabel classifiers with a probabilistic
  thresholding strategy}.
\newblock \bibinfo{journal}{\emph{Pattern Recognition}} \bibinfo{volume}{45},
  \bibinfo{number}{2} (\bibinfo{year}{2012}), \bibinfo{pages}{876--883}.
\newblock


\bibitem[{Read} et~al\mbox{.}(2011)]%
        {read2011classifier}
\bibfield{author}{\bibinfo{person}{Jesse {Read}}, \bibinfo{person}{Bernhard
  {Pfahringer}}, \bibinfo{person}{Geoff {Holmes}}, {and} \bibinfo{person}{Eibe
  {Frank}}.} \bibinfo{year}{2011}\natexlab{}.
\newblock \showarticletitle{Classifier chains for multi-label classification}.
\newblock \bibinfo{journal}{\emph{Machine Learning}} \bibinfo{volume}{85},
  \bibinfo{number}{3} (\bibinfo{year}{2011}), \bibinfo{pages}{333--359}.
\newblock


\bibitem[{Simonyan} and {Zisserman}(2015)]%
        {simonyan2015very}
\bibfield{author}{\bibinfo{person}{Karen {Simonyan}} {and}
  \bibinfo{person}{Andrew {Zisserman}}.} \bibinfo{year}{2015}\natexlab{}.
\newblock \showarticletitle{Very Deep Convolutional Networks for Large-Scale
  Image Recognition}. In \bibinfo{booktitle}{\emph{ICLR 2015 : International
  Conference on Learning Representations 2015}}.
\newblock


\bibitem[{Tang} et~al\mbox{.}(2015)]%
        {tang2015line}
\bibfield{author}{\bibinfo{person}{Jian {Tang}}, \bibinfo{person}{Meng {Qu}},
  \bibinfo{person}{Mingzhe {Wang}}, \bibinfo{person}{Ming {Zhang}},
  \bibinfo{person}{Jun {Yan}}, {and} \bibinfo{person}{Qiaozhu {Mei}}.}
  \bibinfo{year}{2015}\natexlab{}.
\newblock \showarticletitle{LINE: Large-scale Information Network Embedding}.
  In \bibinfo{booktitle}{\emph{Proceedings of the 24th International Conference
  on World Wide Web}}. \bibinfo{pages}{1067--1077}.
\newblock


\bibitem[{Vaswani} et~al\mbox{.}(2017)]%
        {vaswani2017attention}
\bibfield{author}{\bibinfo{person}{Ashish {Vaswani}}, \bibinfo{person}{Noam
  {Shazeer}}, \bibinfo{person}{Niki {Parmar}}, \bibinfo{person}{Jakob
  {Uszkoreit}}, \bibinfo{person}{Llion {Jones}}, \bibinfo{person}{Aidan~N.
  {Gomez}}, \bibinfo{person}{Lukasz {Kaiser}}, {and} \bibinfo{person}{Illia
  {Polosukhin}}.} \bibinfo{year}{2017}\natexlab{}.
\newblock \showarticletitle{Attention is All You Need}. In
  \bibinfo{booktitle}{\emph{Proceedings of the 31st International Conference on
  Neural Information Processing Systems}}, Vol.~\bibinfo{volume}{30}.
  \bibinfo{pages}{5998--6008}.
\newblock


\bibitem[{Veličković} et~al\mbox{.}(2018)]%
        {veli2018graph}
\bibfield{author}{\bibinfo{person}{Petar {Veličković}},
  \bibinfo{person}{Guillem {Cucurull}}, \bibinfo{person}{Arantxa {Casanova}},
  \bibinfo{person}{Adriana {Romero}}, \bibinfo{person}{Pietro {Liò}}, {and}
  \bibinfo{person}{Yoshua {Bengio}}.} \bibinfo{year}{2018}\natexlab{}.
\newblock \showarticletitle{Graph Attention Networks}. In
  \bibinfo{booktitle}{\emph{International Conference on Learning
  Representations}}.
\newblock


\bibitem[{Vu} et~al\mbox{.}(2020)]%
        {vu2020privacy}
\bibfield{author}{\bibinfo{person}{Xuan-Son {Vu}}, \bibinfo{person}{Duc-Trong
  {Le}}, \bibinfo{person}{Christoffer {Edlund}}, \bibinfo{person}{Lili
  {Jiang}}, {and} \bibinfo{person}{Hoang~D. {Nguyen}}.}
  \bibinfo{year}{2020}\natexlab{}.
\newblock \showarticletitle{Privacy-Preserving Visual Content Tagging using
  Graph Transformer Networks}. In \bibinfo{booktitle}{\emph{Proceedings of the
  28th ACM International Conference on Multimedia}}.
  \bibinfo{pages}{2299--2307}.
\newblock


\bibitem[{Wang} et~al\mbox{.}(2016)]%
        {wang2016cnn}
\bibfield{author}{\bibinfo{person}{Jiang {Wang}}, \bibinfo{person}{Yi {Yang}},
  \bibinfo{person}{Junhua {Mao}}, \bibinfo{person}{Zhiheng {Huang}},
  \bibinfo{person}{Chang {Huang}}, {and} \bibinfo{person}{Wei {Xu}}.}
  \bibinfo{year}{2016}\natexlab{}.
\newblock \showarticletitle{CNN-RNN: A Unified Framework for Multi-label Image
  Classification}. In \bibinfo{booktitle}{\emph{2016 IEEE Conference on
  Computer Vision and Pattern Recognition (CVPR)}}.
  \bibinfo{pages}{2285--2294}.
\newblock


\bibitem[{Wang} et~al\mbox{.}(2020)]%
        {wang2020multi}
\bibfield{author}{\bibinfo{person}{Ya {Wang}}, \bibinfo{person}{Dongliang
  {He}}, \bibinfo{person}{Fu {Li}}, \bibinfo{person}{Xiang {Long}},
  \bibinfo{person}{Zhichao {Zhou}}, \bibinfo{person}{Jinwen {Ma}}, {and}
  \bibinfo{person}{Shilei {Wen}}.} \bibinfo{year}{2020}\natexlab{}.
\newblock \showarticletitle{Multi-Label Classification with Label Graph
  Superimposing}. In \bibinfo{booktitle}{\emph{Proceedings of the AAAI
  Conference on Artificial Intelligence}}, Vol.~\bibinfo{volume}{34}.
  \bibinfo{pages}{12265--12272}.
\newblock


\bibitem[{Wang} et~al\mbox{.}(2017)]%
        {wang2017multi}
\bibfield{author}{\bibinfo{person}{Zhouxia {Wang}}, \bibinfo{person}{Tianshui
  {Chen}}, \bibinfo{person}{Guanbin {Li}}, \bibinfo{person}{Ruijia {Xu}}, {and}
  \bibinfo{person}{Liang {Lin}}.} \bibinfo{year}{2017}\natexlab{}.
\newblock \showarticletitle{Multi-label Image Recognition by Recurrently
  Discovering Attentional Regions}. In \bibinfo{booktitle}{\emph{2017 IEEE
  International Conference on Computer Vision (ICCV)}}.
  \bibinfo{pages}{464--472}.
\newblock


\bibitem[{Wei} et~al\mbox{.}(2016)]%
        {wei2016hcp}
\bibfield{author}{\bibinfo{person}{Yunchao {Wei}}, \bibinfo{person}{Wei {Xia}},
  \bibinfo{person}{Min {Lin}}, \bibinfo{person}{Junshi {Huang}},
  \bibinfo{person}{Bingbing {Ni}}, \bibinfo{person}{Jian {Dong}},
  \bibinfo{person}{Yao {Zhao}}, {and} \bibinfo{person}{Shuicheng {Yan}}.}
  \bibinfo{year}{2016}\natexlab{}.
\newblock \showarticletitle{HCP: A Flexible CNN Framework for Multi-Label Image
  Classification}.
\newblock \bibinfo{journal}{\emph{IEEE Transactions on Pattern Analysis and
  Machine Intelligence}} \bibinfo{volume}{38}, \bibinfo{number}{9}
  (\bibinfo{year}{2016}), \bibinfo{pages}{1901--1907}.
\newblock


\bibitem[{Xie} et~al\mbox{.}(2017)]%
        {xie2017aggregated}
\bibfield{author}{\bibinfo{person}{Saining {Xie}}, \bibinfo{person}{Ross
  {Girshick}}, \bibinfo{person}{Piotr {Dollar}}, \bibinfo{person}{Zhuowen
  {Tu}}, {and} \bibinfo{person}{Kaiming {He}}.}
  \bibinfo{year}{2017}\natexlab{}.
\newblock \showarticletitle{Aggregated Residual Transformations for Deep Neural
  Networks}. In \bibinfo{booktitle}{\emph{2017 IEEE Conference on Computer
  Vision and Pattern Recognition (CVPR)}}. \bibinfo{pages}{5987--5995}.
\newblock


\bibitem[{Yalniz} et~al\mbox{.}(2019)]%
        {yalniz2019billion}
\bibfield{author}{\bibinfo{person}{I.~Zeki {Yalniz}}, \bibinfo{person}{Hervé
  {Jégou}}, \bibinfo{person}{Kan {Chen}}, \bibinfo{person}{Manohar {Paluri}},
  {and} \bibinfo{person}{Dhruv {Mahajan}}.} \bibinfo{year}{2019}\natexlab{}.
\newblock \showarticletitle{Billion-scale semi-supervised learning for image
  classification.}
\newblock \bibinfo{journal}{\emph{arXiv preprint arXiv:1905.00546}}
  (\bibinfo{year}{2019}).
\newblock


\bibitem[{Yang} et~al\mbox{.}(2016)]%
        {yang2016exploit}
\bibfield{author}{\bibinfo{person}{Hao {Yang}}, \bibinfo{person}{Joey~Tianyi
  {Zhou}}, \bibinfo{person}{Yu {Zhang}}, \bibinfo{person}{Bin-Bin {Gao}},
  \bibinfo{person}{Jianxin {Wu}}, {and} \bibinfo{person}{Jianfei {Cai}}.}
  \bibinfo{year}{2016}\natexlab{}.
\newblock \showarticletitle{Exploit Bounding Box Annotations for Multi-Label
  Object Recognition}. In \bibinfo{booktitle}{\emph{2016 IEEE Conference on
  Computer Vision and Pattern Recognition (CVPR)}}. \bibinfo{pages}{280--288}.
\newblock


\bibitem[{Yazici} et~al\mbox{.}(2020)]%
        {yazici2020orderless}
\bibfield{author}{\bibinfo{person}{Vacit~Oguz {Yazici}}, \bibinfo{person}{Abel
  {Gonzalez-Garcia}}, \bibinfo{person}{Arnau {Ramisa}},
  \bibinfo{person}{Bartlomiej {Twardowski}}, {and} \bibinfo{person}{Joost
  van~de {Weijer}}.} \bibinfo{year}{2020}\natexlab{}.
\newblock \showarticletitle{Orderless Recurrent Models for Multi-Label
  Classification}. In \bibinfo{booktitle}{\emph{2020 IEEE/CVF Conference on
  Computer Vision and Pattern Recognition (CVPR)}}.
  \bibinfo{pages}{13440--13449}.
\newblock


\bibitem[{Yun} et~al\mbox{.}(2019)]%
        {yun2019graph}
\bibfield{author}{\bibinfo{person}{Seongjun {Yun}}, \bibinfo{person}{Minbyul
  {Jeong}}, \bibinfo{person}{Raehyun {Kim}}, \bibinfo{person}{Jaewoo {Kang}},
  {and} \bibinfo{person}{Hyunwoo~J. {Kim}}.} \bibinfo{year}{2019}\natexlab{}.
\newblock \showarticletitle{Graph Transformer Networks}. In
  \bibinfo{booktitle}{\emph{33rd Annual Conference on Neural Information
  Processing Systems, NeurIPS 2019}}, Vol.~\bibinfo{volume}{32}.
  \bibinfo{pages}{11960--11970}.
\newblock


\bibitem[{Zhao} et~al\mbox{.}(2016)]%
        {zhao2016deep}
\bibfield{author}{\bibinfo{person}{Kaili {Zhao}}, \bibinfo{person}{Wen-Sheng
  {Chu}}, {and} \bibinfo{person}{Honggang {Zhang}}.}
  \bibinfo{year}{2016}\natexlab{}.
\newblock \showarticletitle{Deep Region and Multi-label Learning for Facial
  Action Unit Detection}. In \bibinfo{booktitle}{\emph{2016 IEEE Conference on
  Computer Vision and Pattern Recognition (CVPR)}}.
  \bibinfo{pages}{3391--3399}.
\newblock


\bibitem[{Zhu} et~al\mbox{.}(2017)]%
        {zhu2017learning}
\bibfield{author}{\bibinfo{person}{Feng {Zhu}}, \bibinfo{person}{Hongsheng
  {Li}}, \bibinfo{person}{Wanli {Ouyang}}, \bibinfo{person}{Nenghai {Yu}},
  {and} \bibinfo{person}{Xiaogang {Wang}}.} \bibinfo{year}{2017}\natexlab{}.
\newblock \showarticletitle{Learning Spatial Regularization with Image-Level
  Supervisions for Multi-label Image Classification}. In
  \bibinfo{booktitle}{\emph{2017 IEEE Conference on Computer Vision and Pattern
  Recognition (CVPR)}}. \bibinfo{pages}{2027--2036}.
\newblock


\bibitem[{Zhuang} et~al\mbox{.}(2018)]%
        {zhuang2018multi}
\bibfield{author}{\bibinfo{person}{Ni {Zhuang}}, \bibinfo{person}{Yan {Yan}},
  \bibinfo{person}{Si {Chen}}, \bibinfo{person}{Hanzi {Wang}}, {and}
  \bibinfo{person}{Chunhua {Shen}}.} \bibinfo{year}{2018}\natexlab{}.
\newblock \showarticletitle{Multi-label learning based deep transfer neural
  network for facial attribute classification}.
\newblock \bibinfo{journal}{\emph{Pattern Recognition}}  \bibinfo{volume}{80}
  (\bibinfo{year}{2018}), \bibinfo{pages}{225--240}.
\newblock


\end{thebibliography}
\end{document}